%% file: ijcai25.tex
\documentclass{article}
\usepackage{ijcai25}

\usepackage{times}
\usepackage{soul}
\usepackage{url}
\usepackage[hidelinks]{hyperref}
\usepackage[utf8]{inputenc}
\usepackage[small]{caption}
\usepackage{graphicx}
\usepackage{amsmath}
\usepackage{amsthm}
\usepackage{booktabs}
\usepackage{algorithm}
\usepackage{algorithmic}
\usepackage[switch]{lineno}

\usepackage{bm}
\usepackage{multirow}
\usepackage{tabularx}
\usepackage{caption}
\usepackage{subcaption}

\usepackage{amssymb}
\usepackage{balance}

\title{RAMer: Reconstruction-based Adversarial Model for Multi-party Multi-modal Multi-label Emotion Recognition}

\author{
Xudong Yang$^1$,
Yizhang Zhu$^1$,
Hanfeng Liu$^1$,
Zeyi Wen$^{1,2}$,
Nan Tang$^{1,2}$ \And
Yuyu Luo$^{1,2}$ \\
\affiliations
$^1$The Hong Kong University of Science and Technology (Guangzhou), Guangzhou, China \\
$^2$The Hong Kong University of Science and Technology, Hong Kong SAR, China\\
}

\begin{document}

\maketitle

\begin{abstract}
Conventional Multi-modal multi-label emotion recognition (MMER) assumes complete access to visual, textual, and acoustic modalities. However, real-world \textit{multi-party} settings often violate this assumption, as non-speakers frequently lack acoustic and textual inputs, leading to a significant degradation in model performance. Existing approaches also tend to unify heterogeneous modalities into a single representation, overlooking each modality’s unique characteristics.
To address these challenges, we propose \textbf{RAMer} (\textbf{R}econstruction-based \textbf{A}dversarial \textbf{M}odel for \textbf{E}motion \textbf{R}ecognition), which refines multi-modal representations by not only exploring modality commonality and specificity but crucially by leveraging reconstructed features, enhanced by contrastive learning, to overcome data incompleteness and enrich feature quality. \textbf{RAMer} also introduces a personality auxiliary task to complement missing modalities using modality-level attention, improving emotion reasoning. To further strengthen the model's ability to capture label and modality interdependency, we propose a stack shuffle strategy to enrich correlations between labels and modality-specific features.
Experiments on three benchmarks, i.e., MEmoR, CMU-MOSEI, and $M^3$ED, demonstrate that \textbf{RAMer} achieves state-of-the-art performance in dyadic and multi-party MMER scenarios.
\end{abstract}

\input{source/introduction}

\input{source/related_work}
\input{source/methodology}
\input{source/experimets}
\input{source/conclusion}

\bibliographystyle{named}
\bibliography{ijcai25}

\end{document}

%% file: source/introduction.tex
\section{Introduction}

\begin{figure}[t] 
    \centering 
    \includegraphics[width=\columnwidth]{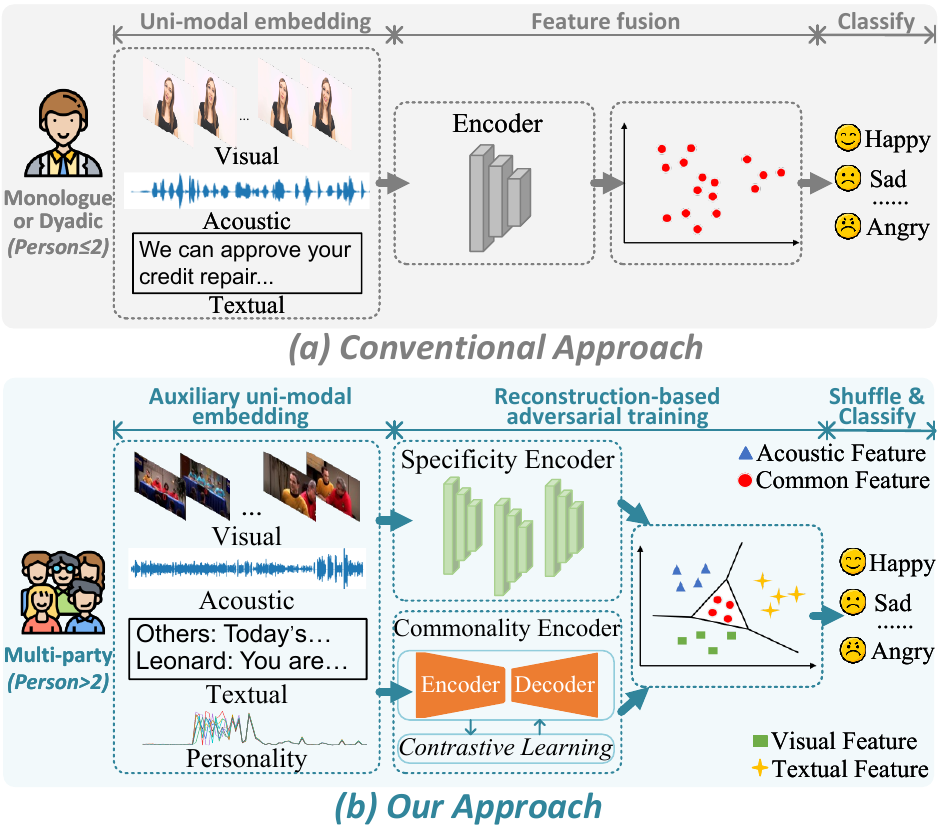}
    \vspace{-1.5em}
    \caption{(a) shows the conventional approach for MMER in monologue and dyadic conversations with complete modalities in a uniform representation; (b) depicts our approach for multi-party conversations with incomplete modalities, reconstructing and projecting them into both specificity and commonality representations.}
    \vspace{-0.7em}
    \label{fig: motivation}
\end{figure}

Emotion recognition from videos is crucial for advancing human-computer interaction and social intelligence. Multi-modal multi-label emotion recognition (MMER) leverages visual, textual, and acoustic signals to identify multiple emotions (e.g., happy, sad) simultaneously~\cite{MMER1,MMER2}.
Conventional MMER methods, as shown in Figure~\ref{fig: motivation}(a), typically focus on monologue or dyadic settings, assuming all modalities are fully available. However, real-world conversations often involve multiple participants (i.e., \textit{multi-party}) with incomplete modality data for non-speakers who always lack acoustic and textual signals.

\textbf{\textit{Multi-party} MMER}, a more complex and practical setting, introduces three key challenges. 
Firstly, handling incomplete modalities is a significant challenge, which requires robust methods to reconstruct or infer missing information. Most existing works~\cite{MMER2,Tailor,adversarial_masking} assume complete modality access and encode each modality independently, overlooking missing data. While some methods ~\cite{Dialoguegcn,DialogueCRN} leverage speaker-aware context modeling, their performance degrades in multi-party settings where non-speakers often lack critical modalities. 
Secondly, representing diverse modalities effectively remains challenging, often requiring techniques that can not only integrate disparate information but also reconstruct rich, complete representations from potential modalities. Current fusion strategies, such as aggregation-based methods (e.g., concatenation, averaging)\cite{MEmoR} and hybrid approaches~\cite{multimodal_survey2}, project modalities into a shared subspace, often neglecting their unique characteristics and reducing discriminative ability. Recent methods~\cite{Tailor} attempt to separate modality-specific and shared features but often suffer from information loss due to inadequate handling of inter-modal correlations. Similarly, methods preserving modality-specific information~\cite{CARAT} may overlook cross-modal commonalities, limiting their ability to fully capture inter-modal relationships.
Finally, multi-label learning presents challenges in modeling robust label correlations and capturing complex interdependency between modalities and labels. Existing approaches~\cite{bloom_filters,ma2021label} often fail to fully exploit collaborative label relationships. Moreover, emotions vary across modalities, and different emotions rely on distinct modality features, further complicating the task.

To address these issues, we propose \textbf{RAMer}, a novel framework designed to tackle the challenges of the \textbf{\textit{Multi-party} MMER} problem. RAMer integrates multimodal representation learning with multi-label modeling to effectively handle incomplete modalities in multi-party settings.

As illustrated in Figure~\ref{fig: motivation}(b), RAMer addresses the challenge of multi-party MMER by following techniques.
To address the challenge of incomplete modalities, we propose an auxiliary task that incorporates external knowledge, such as personality traits, to complement the existing modalities. Leveraging this, we employ modality-level attention mechanisms to capture both inter- and intra-personal features. A reconstruction-based network is utilized to recover and enrich the features of any modality by leveraging information from the other modalities.

To represent diverse modalities effectively and capture discriminative features, we design an adversarial network that extracts commonality across modalities while amplifying the specificity inherent to each one. This helps ensure minimal information loss during the fusion process. 

Additionally, to model robust interconnections between modalities and labels, we propose a novel modality shuffle strategy. This strategy enriches the feature space by shuffling both samples and modalities, based on the commonality and specificity of the modalities, improving the model's ability to capture label correlations and modality-to-label relationships.

In summary, the contributions of this work are:
\begin{itemize}
  \item 
  \textit{A Novel Model for the Multi-party MMER Problem.} We present \textbf{RAMer}, a new framework that centrally integrates feature reconstruction within an adversarial learning paradigm. \textbf{RAMer} adeptly captures both commonality and specificity across modalities, crucially utilizing robustly reconstructed features to significantly improve emotion recognition, especially even with incomplete modality data. 
  \item \textit{Optimization Techniques.} To enhance the robustness of multi-party emotion recognition, \textbf{RAMer} employs contrastive learning to enrich reconstructed features and integrates a personality auxiliary task to capture modality-level attention. We also propose a stack shuffle strategy, enhancing the modeling of label correlations and modality-to-label relationships by leveraging the commonality and specificity of different modalities.
  \item \textit{Extensive Experiments.} We conduct comprehensive experiments on three benchmarks, i.e., MEmoR, CMU-MOSEI, and $M^3$ED, across various conversation scenarios. Results show that \textbf{RAMer} surpasses existing approaches and achieves state-of-the-art performance in both dyadic and multi-party MMER problems.
\end{itemize}

%% file: source/related_work.tex
\section{Related Work}

\paragraph{Multi-modal Representation Learning.}
Emotion recognition has progressed from uni-modal approaches~\cite{huang2021audio,saha2020towards} to multi-modal methods~\cite{lv2021progressive,Tailor} that exploit complementary features across modalities. While uni-modal approaches often face recognition biases~\cite{huang2021audio}, multi-modal learning has gained significant attention, with a key challenge being the effective integration of heterogeneous modalities. Early fusion methods, such as concatenation~\cite{ngiam2011multimodal}, tensor fusion~\cite{liu2018efficient}, and averaging~\cite{fusenet}, struggle with modality gaps that hinder effective feature alignment. To address this, attention-based methods~\cite{adversarial_masking,cross_adaptation} leverage cross-attention mechanisms to dynamically align features in the latent space, while contrastive learning~\cite{chen2020simple_CL,CARAT} further improves robustness. However, most attention-based methods merge modalities into a joint embedding, often overlooking modality-specific features.

\begin{figure*}[t] 
    \centering 
    \includegraphics[width=0.96\linewidth]{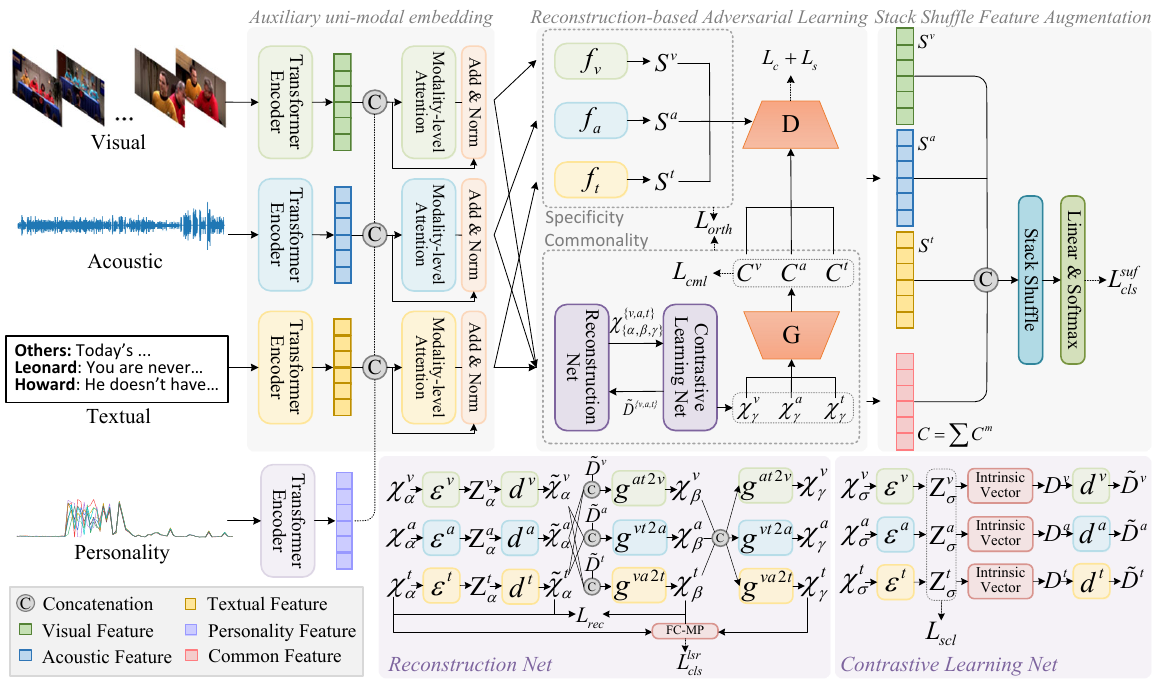}
    \caption{The framework of RAMer. Given incomplete multi-modal inputs, RAMer first encodes each individual modality through an auxiliary task, then feeds the features into a reconstruction-based adversarial network to extract specificity and commonality. Finally, a stack shuffle strategy is employed to learn enhanced representations.}
    \label{framework}
    \vspace{-0.2em}
\end{figure*}

\paragraph{Multi-label Emotion Recognition in Videos.}
MMER involves assigning multiple emotion labels to a target person in the video. Early methods treated multi-label classification as independent binary tasks~\cite{boutell2004learning}, but advancements explore label correlations using techniques like Adjacency-based Similarity Graph Embedding(ASGE)~\cite{ASGE}, Graph Convolutional Network (GCN)~\cite{chen2019multi}, and multi-task pattern~\cite{tsai2020order} to explore label correlations. Some noteworthy strategies~\cite {zhang2021bilabel,Tailor} focus on modeling label-feature correlations through label-specific representations enabled by visual attention~\cite{chen2019multi} and transformers~\cite{Tailor}. Beyond monologue settings, MMER in conversations has gained interest, using GCN~\cite{Dialoguegcn} and memory networks~\cite{memory_network} to model dynamic speaker interactions. However, multi-party MMER is especially challenging, as it requires recognizing emotions for multiple speakers with incomplete modalities. Additionally, the influence of individual personalities on label-feature correlations remains underexplored.

\paragraph{Adversarial Training.}
Adversarial training (AT)~\cite{GAN}, involves two models: a discriminator that estimates the probability of samples, and a generator that creates samples indistinguishable from actual data. This setup forms a minimax two-player game, enhancing the robustness of the model. The technique has since been adapted for CV and NLP applications~\cite{GAN_survey}. For instance, Miyato et al.~\cite{miyato2016adversarial} extended AT to text categorization by introducing perturbations to word embeddings. Wu et al.~\cite{adversarial_relation_extraction} applied it within a multi-label learning framework to facilitate relationship extraction. Additionally, AT has been used to learn joint distributions between multi-modal~\cite{tsai2018learning}. More recently, Ge et al.~\cite{adversarial_masking} applied AT to reduce modal and data biases in MMER tasks. However, Zhang et al.~\cite{Tailor} implemented AT to extract multi-modal commonality and diversity, but suffered a significant loss of modality information due to inadequate cross-modal information fusion.

%% file: source/methodology.tex
\section{Problem Formulation}
\label{gen_inst}

In this section, we introduce the notations used and formally define the \textit{Multi-party} Multi-modal Multi-label Emotion Recognition (\textit{Multi-party} MMER) problem.

\paragraph{Notations.}
We use lowercase letters for scalars (e.g., $v$), uppercase letters for vectors (e.g., $Y$), and boldface for matrices (e.g., $\bm{X}$).
A data sample is represented by the tuple $\left( V, P_{t}, S_{r}, E_{t,r} \right)$, where:

\begin{itemize}
    \item $V = \left( \left\{ P_i \right\}_{i=1}^T, \left\{ S_j \right\}_{j=1}^R \right)$ is a video clip containing $T$ persons and $R$ semantic segments.
    \item $\left\{ P_i \right\}_{i=1}^T$ refers to the set of target persons, and $\left\{ S_j \right\}_{j=1}^R$ represents the target segments, each annotated with an emotion moment.
    \item $E_{t,r}$ denotes the labeled emotion for person $P_{t}$ in  $S_{r}$.
\end{itemize}

Each sample involves modalities such as visual ($v$), acoustic ($a$), textual ($t$), and personality traits ($p$).

For each modality $m \in \left\{ v, a, t, p \right\}$, the corresponding features are represented as $\left( \bm{\mathcal X}^{1}, \bm{\mathcal X}^{2}, \cdots, \bm{\mathcal X}^{m} \right)$, where $\bm{\mathcal X}^{k} \in \mathbb{R}^{l_k \times d_k}$ represents the feature space of the $k$-th modality. Here:
$l_k$ denotes the sequence length, and
$d_k$ denotes the dimension of the modality.
Let $\mathcal{Y} = \left\{ y_{1}, y_{2}, \cdots, y_{\zeta} \right\}$ represent a label space with $\zeta$ possible emotion labels.



\paragraph{Multi-party MMER Problem.}
Given a training dataset $\mathcal{D} = \left\{ \bm{X}_{\tau}^{\left\{ 1, 2, \cdots, m \right\}}, Y_{\tau} \right\}_{\tau=1}^{N}$ with $N$ data samples, where:
(1) $\bm{X}_{\tau}^{m} \in \bm{\mathcal{X}}^{m}$ represents the features for each modality $m$ in sample $\tau$, and
(2) $Y_{\tau} = \left\{ 0, 1 \right\}^{\zeta}$ is a multi-hot vector indicating the presence ($1$) or absence ($0$) of emotion labels, where $Y_{\tau}^{\upsilon} = 1$ indicates that sample $\tau$ belongs to class $\upsilon$, and $Y_{\tau}^{\upsilon} = 0$ otherwise.
%
%
The \textbf{goal} of the Multi-party MMER problem is to learn a function $\mathcal{F}: \bm{\mathcal{X}}^{1} \times \bm{\mathcal{X}}^{2} \times \cdots \times \bm{\mathcal{X}}^{m} \mapsto \mathcal{Y}$ that predicts the target emotion $E_{t,r}$ for person $P_{t}$ in segment $S_{r}$, leveraging contextual information from multiple modalities.


\paragraph{Discussion.}
It is important to note that the target person $P_{t}$ may have incomplete modality information, meaning they may not simultaneously possess visual, textual, or acoustic representations. This introduces uncertainty in the modality of the target segment $S_{r}$, making the prediction task more challenging.

\section{Methodology}

Figure~\ref{framework} shows the framework of RAMer, which consists of three components: auxiliary uni-modal embedding, reconstruction-based adversarial Learning, and stack shuffle feature augmentation.

\subsection{Auxiliary Uni-modal Embedding}
\label{uni_modal_freature_ext}

 To extract contextual information from each modality, we employ four independent transformer encoders~\cite{transformer}, each dedicated to a specific modality $m$.Each encoder consists of $n_m$ identical layers to ensure consistent and deep representation. For multi-party conversation videos with $T$ participants with incomplete modalities, we introduce an optional auxiliary task leveraging personality to complement missing modalities. 
 Specifically, we concatenate personality embedding $\bm{\mathcal X}^{p}$ with each modality $\bm{\mathcal X}^{m} \in \left\{v,t,a \right\}$ to enrich the feature space. We then apply the scaled dot-product attention to compute inter-person attention across the person dimension within each segment, and intra-person attention along the segment dimension for each individual~\cite{MEmoR}. This modality-level attention mechanism is designed to enhance the model's emotion reasoning ability by effectively capturing both interpersonal dynamics and temporal patterns within the data.
In this way, we obtain personality enhanced representation $\bm{\mathcal X}_{\alpha}^{m} \in \mathbb{R}^{l\times d}$.

\subsection{Reconstruction-based Adversarial Learning}
\label{Reconstruction_based_Adversarial}

The second component leverages multiple modalities by capturing inter-modal commonalities while preserving modality-specific features. To address the limitations of adversarial networks~\cite{GAN,Tailor}, which can result in information loss and difficulty in learning modality-label dependencies, we introduce a reconstruction-based approach. It employs contrastive learning to learn modality-independent but label-relevant representations while reconstructing missing modalities during training to enhance robustness.

\paragraph{Adversarial Training.} To balance modality specificity and commonality, we adopt adversarial training to extract discriminative features. The uni-modal embeddings $\bm{\mathcal X}_{\alpha}^{m}$ are fed to three fully connected networks $f_m$ to extract specificity $\bm{S}^{m}, m \in \left\{ v, a, t\right\}$. In parallel, $\bm{\mathcal X}_{\alpha}^{m}$ are also passed through a reconstruction network, which is coupled with a contrastive learning network, followed by a generator $ G\left ( \cdot;\theta_{G} \right )$ to derive the commonality $\bm{C}^{m}$. Both specificity and commonality are then passed through linear layers with softmax activation in the discriminator $ D\left ( \cdot;\theta_{D} \right )$ that is designed to distinguish which modality the inputs come from. The generator captures commonality ${C}^{m}$ by projecting different reconstructed embedding $\bm{\mathcal{X}}_{\gamma }^{m}$ into a shared latent subspace, ensuring distributional alignment across modalities.
 Consequently, this architecture encourages the generator $G(\cdot; \theta_{G})$ to produce outputs that challenge the discriminator $ D\left ( \cdot;\theta_{D} \right )$ by obscuring the source modality of $\bm{C}^{m}$. The generator and discriminator are jointly trained in a game-theoretic setup to enhance feature robustness against modality-specific biases. Both the commonality adversarial loss $\mathcal{L}_C$ and the specificity adversarial loss $\mathcal{L}_S$ are calculated by cross-entropy loss as,
\vspace{-1.0em}
\begin{equation}
\small
\mathcal{L}_C = -\frac{1}{N} \sum_{m \in \{v,t,a\}} \sum_{\tau=1}^{N} \Big(U^{m} \log \big(D(C_{\tau}^m ; \theta_D)\big)\Big),
\end{equation}
\vspace{-1.0em}
\begin{equation}
\small
\mathcal{L}_S = -\frac{1}{N} \sum_{m \in \{v,t,a\}} \sum_{\tau=1}^{N} \Big(U^{m} \log \big(D(S_{\tau}^m ; \theta_D)\big)\Big),
\end{equation}
where $U \in \left\{ U^v,U^t,U^a\right\}$ represents the ground truth label corresponding to the discriminator's input.


In the shared subspace, it is advantageous to employ a unified representation of various modalities to facilitate multi-label classification. This representation is designed to eliminate redundant information and extract the elements common to the different modalities, thereby introducing a common semantic loss defined as,
\vspace{-0.8em}
\begin{equation}
\small
\label{L_cml}
\mathcal{L}_{cml} = - \sum_{m \in \{v, t, a\}} \sum_{\tau=1}^{N} \sum_{\upsilon=1}^{\zeta} y_{\tau}^{\upsilon} \log \hat{y}_{\tau}^{\upsilon,m} + (1 - y_{\tau}^{\upsilon}) \log(1 - \hat{y}_{\tau}^{\upsilon,m}),
\end{equation}
where $\hat{y}_{\tau}^{\upsilon,m}$ is predicted with $\bm{C}_{m}$ and $y_{\tau}^{\upsilon}$ is the ground-truth label.
To encode diverse aspects of multi-modal data, we introduce an orthogonal loss $\mathcal{L}_{orth}$ that encourages the commonality $\bm{C}^{m}$ and specificity $\bm{S}^{m}$ subspaces to remain distinct by minimizing their overlap.
\vspace{-1.0em}
\begin{equation}
\small
\label{L_orth}
\mathcal{L}_{orth} = - \sum_{m \in \{v, t, a\}} \sum_{\tau=1}^{N} \left\| (\bm{C}_{\tau}^{m})^{T} \bm{S}_{\tau}^{m} \right\|_{F}^{2},
\end{equation}
where $\left\| \cdot \right\|_{F}$ is Frobenius norm. Hereby, the objective of adversarial training $\mathcal{L}_{adv}$ is
\begin{equation}
\small
\label{L_adv}
    \mathcal{L}_{adv} = \lambda_a\left ( \mathcal{L}_{C} + \mathcal{L}_{S}\right )+ \lambda_o \mathcal{L}_{orth}+ \lambda_c \mathcal{L}_{cml} ,
\end{equation}
where $\lambda_a,\lambda_o$ and $\lambda_c$ are trade-off parameters.


\paragraph{Multi-modal Feature Reconstruction.} To reconstruct the features of any modality by leveraging information from the other modalities. We employ a reconstruction network that is composed of modality-specific encoders $ \varepsilon^m$, decoders $d^m$, and a two-level reconstruction process utilizing multi-layer linear networks $g(\cdot)$. Given input $\bm{\mathcal X}_{\alpha}^{m}$ from different modality, three encoders $\varepsilon ^m$ that consist of MLPs are utilized to project $\bm{\bm{\mathcal{X}}}_{\alpha }^{m}$ into latent embedding $\bm{Z}_{\alpha }^{m}$ within the latent space $\mathcal{S}^z$. Subsequently, three corresponding decoders $d^m$ transform these latent vectors into the decoded vectors $\bm{\mathcal{\widetilde{X}}}_{\alpha }^{m}$. 
At the first level of reconstruction network, the intrinsic vector $\bm{\widetilde{D}}^{m}$ that derived from contrastive learning network and semantic features $\mathcal{\widetilde{X}}_{\alpha }^{\left\{v,t,a \right\}\backslash m}$ are concatenated to form the input, which is processed to produce $\bm{\mathcal{X}}_{\beta }^{m}$ used for the second-level reconstruction network. Hereby, the reconstruction network can be formulated as,
\vspace{-.4em}
\begin{equation}
    \bm{\bm{\mathcal{X}}}_{\gamma }^{m}=g\left ( g\left ( d^m(\varepsilon^m(\bm{\mathcal X}_{\alpha}^{m};\theta_{m}) ), \bm{\widetilde{D}}^{m} \right ) \right ).
\end{equation}
The obtained three embedding $\bm{\mathcal{X}}_{\alpha }^{m},\bm{\mathcal{X}}_{\beta }^{m}$, and $\bm{\mathcal{X}}_{\gamma }^{m}$ from three distinct feature spaces are fed into fully connected network followed by max pooling.
We can formulate the reconstruction loss $\mathcal{L}_{rec}$ and classification loss $\mathcal{L}_{cls}^{lsr}$ as,
\vspace{-.5em}
\begin{equation}
\small
\label{L_rec}
\mathcal{L}_{rec} = \sum_{m=1}^{M} \left( \left\|\bm{\mathcal{X}}_{\alpha}^{m} - \widetilde{\bm{\mathcal{X}}}_{\alpha}^{m} \right\|_{F} + \left\|\bm{\mathcal{X}}_{\alpha}^{m} - \bm{\mathcal{X}}_{\beta}^{m} \right\|_{F} \right),
\end{equation}
\vspace{-.5em}
\begin{equation}
\small
    \mathcal{L}_{cls}^{lsr}=\lambda_\alpha\mathcal{L}_{B}\left ( \mathcal{S}^\alpha,Y \right )+\lambda_\beta\mathcal{L}_{B}\left ( \mathcal{S}^\beta,Y \right )+\lambda_\gamma\mathcal{L}_{B}\left ( \mathcal{S}^\gamma,Y \right ),
\end{equation}
where $\left\| \cdot \right\|_{F}$ is the Frobenius norm, $\lambda_{\alpha,\beta,\gamma}$ are trade-off parameters, $\mathcal{L}_{B}$ is the binary cross entropy (BCE) loss. 

To capture the feature distributions of different modalities and use them to guide the restoration of incomplete modalities, intrinsic vectors $\bm{\widetilde{D}}^{m}$ obtained through a supervised contrastive learning network~\cite{SCL} are incorporated into the reconstruction network. 
The encoders $\varepsilon^m$ project input $\bm{\mathcal X}_{\sigma}^{m}$ to contrastive embeddings $\bm{Z}_{\sigma}^{m}$, $\sigma \in \left\{\alpha,\beta,\gamma \right\}$. Given a contrastive embedding set $\bm{\mathcal{Z}}=\left\{ \bm{Z}_{\sigma}^{\left\{ v, t, a\right\}} \right\}$, an anchor vector $z_i \in \mathcal{Z}$ and assuming the prototype vector updated during the training process based on the moving average is $\mu_{j,k}^{m}$, where modality $m \in \left\{v,t,a \right\}$, label category $j \in \left [ \zeta \right ]$, label polarity $k \in \left\{pos,neg \right\}$, then the intrinsic vector $\widetilde{D}^{m}$ can be derived from:
\begin{equation}
\delta_j^m = \sum_{k}^{\left\{ pos, neg\right\}} o_{j,k}^m \cdot u_{j,k}^m, \quad  o_{j,k}^m = \frac{\exp(z_i \cdot u_{j,k}^m)}{\sum_{k'}^{\left\{ pos, neg\right\}} \exp(z_i \cdot u_{j,k'}^m)}
\end{equation}
\begin{equation}
    \widetilde{D}^{m}=d^m\left ( \left [ \delta_{1}^{m}, \cdots, \delta_{\zeta}^{m}  \right ]; \theta_m \right ).
\end{equation}
The loss of contrastive learning network is defined as, 
\begin{equation}
\small
\label{L_scl}
\mathcal{L}_{scl}\left ( i, \mathcal{Z} \right ) = \sum_{i \in \mathcal{Z}} -\frac{1}{|\bm{P}(i)|} \sum_{p \in \bm{P}(i)} \log \frac{\exp(z_i \cdot z_p / \eta)}{\sum\limits_{r \in \bm{A}(i)} \exp(z_i \cdot z_r / \eta)},
\end{equation}
where $\bm{P}(i)$ is the positive set, $\eta \in \mathbb{R}^+$ is a temperature parameter, and $\bm{A}(i)= \bm{\mathcal{Z}}\ \backslash\ {\left\{i \right\}}$.

\subsection{Stack Shuffle for Feature Augmentation}
\label{stack_shuffle}

To construct more robust correlations among labels and model the complex interconnections between modalities and labels, we propose a multi-modal feature augmentation strategy that incorporates a stack shuffle mechanism. After obtaining the commonality and specificity representations, we perform sample-wise and modality-wise shuffling processes sequentially on a batch of samples. To strengthen the correlations between labels, we first apply a sample-wise shuffle. The features derived from $\bm{C}^{m}$ and $\bm{S}^{m}$ are split into $k$ stacks along the sample dimension, with the top elements of each stack cyclically popped and appended to form new vectors.  
Next, a modality-wise shuffle is introduced to help the model capture and integrate information across different modalities. For each sample, features are divided into stacks along the modality dimension, and iterative pop-and-append operations are applied.
Finally, the shuffled samples $\mathbf{V}$ are used to fine-tune the classifier $c_{\zeta}$ with the binary cross-entropy (BCE) loss.
\begin{equation}
\small
\label{L_cls_suf}
\mathcal{L}_{cls}^{suf} = -\frac{1}{N} \sum_{m \in \{v,t,a\}} \sum_{\tau=1}^{N} \Big(Y^{m} \log \big(c_{\zeta}(\mathbf{V}^m )\big)\Big),
\end{equation}
Combing the Eq.\eqref{L_adv}, Eq.\eqref{L_rec} $\sim$ Eq.\eqref{L_scl} and Eq.\eqref{L_cls_suf}, the final objective function $\mathcal{L}$ is formulated as,
\begin{equation}
\mathcal{L} = \mathcal{L}_{cls}^{suf} + \mathcal{L}_{cls}^{lsr} + \lambda_r \mathcal{L}_{rec} + \lambda_s \mathcal{L}_{scl} + \mathcal{L}_{adv}
\end{equation}
where $\lambda_r$, $\lambda_s$ are trade-off parameters.



%% file: source/experimets.tex
\section{Experiments}

\subsection{Experimental Setup}

\paragraph{Datasets and Metrics.}
We evaluated RAMer on three benchmark datasets: MEmoR~\cite{MEmoR}, a multi-party conversation dataset that includes personality, and CMU-MOSEI~\cite{Cmu-mosei} and $M^3$ED~\cite{M3ED}, which are dyadic conversation datasets that do not include personality information. The evaluation is conducted under the protocols of these datasets. For CMU-MOSEI and $M^3$ED, we employed four commonly used evaluation metrics: Accuracy (Acc), Micro-F1, Precision (P), and Recall (R). For MEmoR, we followed the benchmark's protocol and used Micro-F1, Macro-F1, and Weighted-F1 metrics. 

\begin{table*}[t!]
\begin{center}
\resizebox{0.95\textwidth}{!} 
{ 
\begin{tabular}{cc|ccc|ccc}
\hline
\hline
\multirow{2}{*}{Methods} 
& \multirow{2}{*}{Modality} 
& \multicolumn{3}{c|}{Primary}       
& \multicolumn{3}{c}{Fine-grained}  \\ 
\cline{3-8} 

&                           
& Micro-F1 & Macro-F1 & Weighted-F1 
& Micro-F1 & Macro-F1 & Weighted-F1 
\\ 
\hline

MDL with Personality     & $v,a,t,p$                   
& 0.429    & 0.317    & 0.423       
& 0.363    & 0.217    & 0.345       
\\

MDAE                    & $v,a,t,p$                   
& 0.421    & 0.303    & 0.410       
& 0.363    & 0.219    & 0.341       
\\

BiLSTM+TFN             & $v,a,t,p$                   
& 0.470    & 0.310    & 0.454       
& 0.366    & 0.207    & 0.350       
\\

BiLSTM+LMF             & $v,a,t,p$                   
&0.449    & 0.294    & 0.432       
& 0.364    & 0.198    & 0.351       
\\

DialogueGCN            & $v,a,t,p$                   
& 0.441    & 0.310    & 0.425       
& 0.373    & 0.229    & 0.373       
\\

AMER w/o Personality     & $v,a,t$                     
& 0.446    & 0.339    & 0.440       
& 0.401    & 0.246    & 0.379       
\\

AMER                & $v,a,t,p$                   
& 0.477    & 0.353    & 0.465       
& 0.419    & 0.262    & 0.400       
\\ 

DialogueCRN     & $v,a,t,p$
& 0.441    & 0.310    & 0.425 
& 0.373    & 0.229    & 0.373 
\\
TAILOR               & $v,a,t,p$
& 0.341 & 0.287 & 0.326 
& 0.303 & 0.069 & \textbf{0.490} 
\\
CARAT              & $v,a,t,p$
& 0.399 & 0.224 & 0.422 
& 0.346 & 0.090 & 0.483 
\\
\hline
\textbf{RAMer}                   & $v,a,t,p$                  
& \textbf{0.499}    & \textbf{0.402}    & \textbf{0.503}       
& \textbf{0.431}    & \textbf{0.299}   & 0.404       
\\ 
\hline
\hline
\end{tabular}
} 
\vspace{-0.2em}
\caption{Performance comparison on the MEmoR dataset under primary and fine-grained settings. With various modality combinations (visual($v$), acoustic($a$), textual($t$), personality($p$)).}
\label{tab:primary} 
\end{center}
\vspace{-1em}
\end{table*}

\begin{table}[t]
\begin{center}
\resizebox{\columnwidth}{!}
{ 
\begin{tabular}{c|cccc|cccc}
    \hline
    \hline

    \multirow{2}{*}{Methods} & \multicolumn{4}{c|}{Aligned}      & \multicolumn{4}{c}{Unaligned}    \\ 
    \cline{2-9}

    & Acc   & P     & R     & Micro-F1 
    & Acc   & P     & R     & Micro-F1 \\ 
    \hline

    CC                    
    & 0.225 & 0.306 & 0.523 & 0.386    
    & 0.235 & 0.320 & 0.550 & 0.404    
    \\

    ML-GCN                   
    & 0.411 & 0.546 & 0.476 & 0.509    
    & 0.437 & 0.573 & 0.482 & 0.524    
    \\
    
    MulT                    
    & 0.445 & 0.619 & 0.465 & 0.531    
    & 0.423 & 0.636 & 0.445 & 0.523    
    \\

    MISA                    
    & 0.43  & 0.453 & \textbf{0.582} & 0.509    
    & 0.398 & 0.371 & \textbf{0.571} & 0.45     
    \\

    MMS2S                    
    & 0.475 & 0.629 & 0.504 & 0.56     
    & 0.447 & 0.619 & 0.462 & 0.529    
    \\  
    
    HHMPN                  
    & 0.459 & 0.602 & 0.496 & 0.556    
    & 0.434 & 0.591 & 0.476 & 0.528    
    \\ 
    
    TAILOR                  
    & 0.488 & 0.641 & 0.512 & 0.569    
    & 0.46  & 0.639 & 0.452 & 0.529    
    \\ 
    
    AMP                      
    & 0.484 & 0.643 & 0.511 & 0.569    
    & 0.462 & 0.642 & 0.459 & 0.535    
    \\  
    
    CARAT                   
    & 0.494 & 0.661 & 0.518 & 0.581    
    & 0.466 & 0.652 & 0.466 & 0.544    
    \\ 
    \hline 
    
    \textbf{RAMer}                   
    & \textbf{0.505} & \textbf{0.668} & 0.551 & \textbf{0.604}    
    & \textbf{0.469} & \textbf{0.660} & 0.486 & \textbf{0.560}    
    \\ 
    \hline
    \hline
\end{tabular}
}
\vspace{-0.2em}
\caption{Performance Comparison on CMU-MOSEI dataset.}
\label{tab:cmu-mosei} 
\end{center}
\vspace{-1em}
\end{table}

\begin{table}[t]
\begin{center}
\small
\begin{tabular}{ccccc}
\hline
\hline
Methods & Acc   & P     & R     & Micro-F1 \\ 
\hline
MMS2S   & 0.645 & 0.813 & 0.737 & 0.773    \\
HHMPN   & 0.648 & 0.816 & 0.743 & 0.778    \\
TAILOR  & 0.647 & 0.814 & 0.739 & 0.775    \\
AMP     & 0.654 & 0.819 & 0.748 & 0.782    \\
CARAT   & 0.664 & 0.824 & 0.755 & 0.788    \\ 
\hline
\textbf{RAMer}  & \textbf{0.665} & \textbf{0.826} & \textbf{0.759} & \textbf{0.791}     \\ 
\hline
\hline
\end{tabular}
\vspace{-0.2em}
\caption{Performance Comparison on the M$^3$ED dataset} 
\label{tab:m3ed} 
\end{center}
\vspace{-1em}
\end{table}

\paragraph{Baselines.}
For the MEmoR dataset, we compare RAMer with multi-party conversation baselines, including MDL, MDAE~\cite{MDAE}, BiLSTM+TFN~\cite{TFN}, BiLSTM+LMF~\cite{LMF}, DialogueGCN~\cite{Dialoguegcn}, DialogueCRN~\cite{DialogueCRN}, and AMER~\cite{MEmoR}. We further assess its robustness against recent dyadic models, including CARAT~\cite{CARAT} and TAILOR~\cite{Tailor}. For the CMU-MOSEI and $M^3$ED datasets, we test three categories of methods. 1) Classic methods. CC~\cite{CC}, which concatenates all available modalities as input for binary classifiers. 2) Deep-based methods. ML-GCN~\cite{chen2019multi}, using Graph Convolutional Networks to map label representations and capture label correlations. 3) Multi-modal multi-label methods. These include MulT~\cite{cross_adaptation} for cross-modal interactions, MISA~\cite{MISA} for learning modality-invariant and modality-specific features, and methods like MMS2S~\cite{MMER1}, HHMPN~\cite{MMER2}, TAILOR~\cite{Tailor}, AMP~\cite{adversarial_masking}, and CARAT~\cite{CARAT}.


\begin{figure*}[thb]
    \centering
    \begin{subfigure}[b]{0.2\textwidth}
        \centering
        \includegraphics[width=\textwidth]{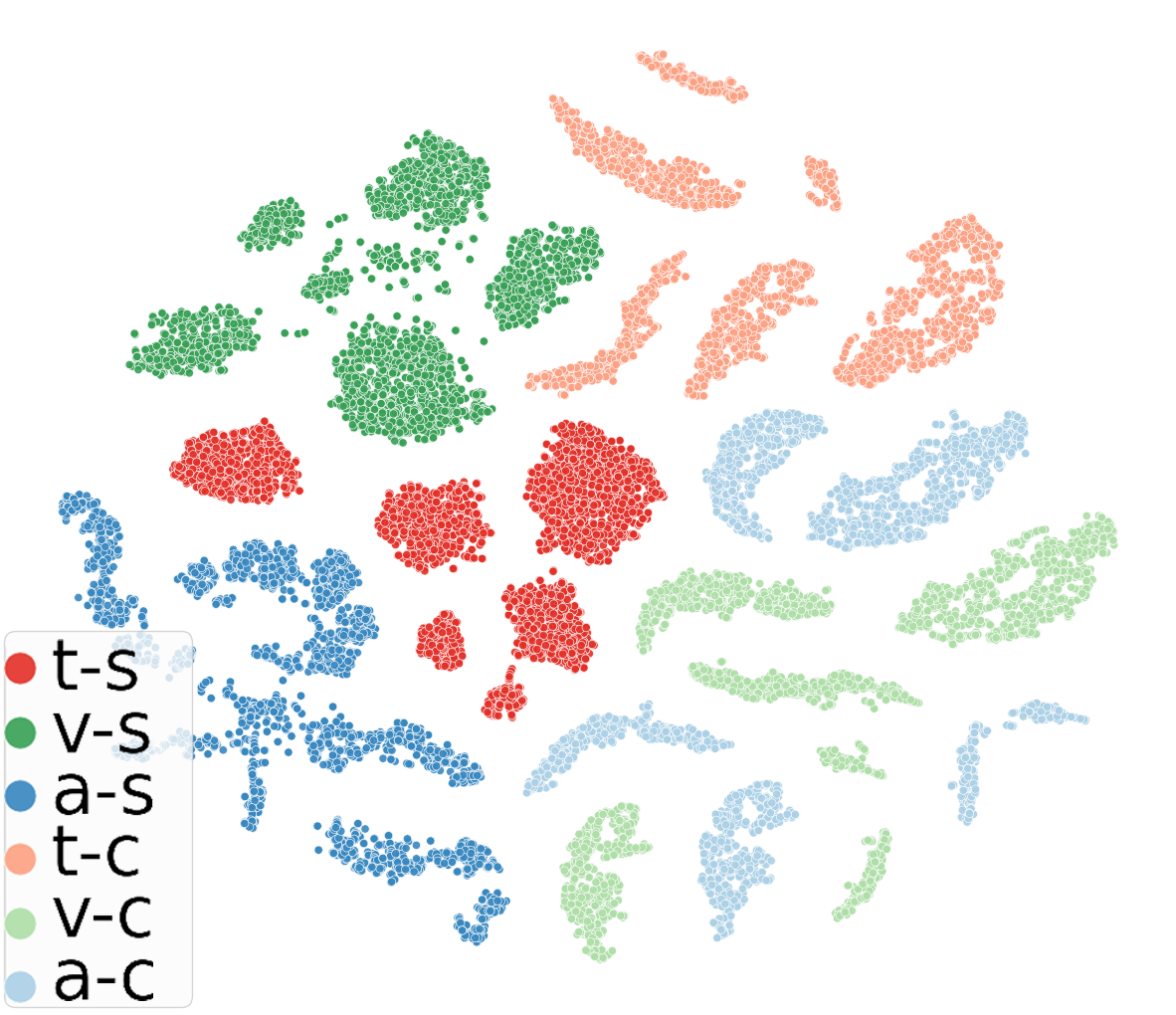}
        \caption{w/o Adversarial Training}
        \label{fig:top-left}
    \end{subfigure}
    \hspace{0.02\textwidth}
    \begin{subfigure}[b]{0.2\textwidth}
        \centering
        \includegraphics[width=\textwidth]{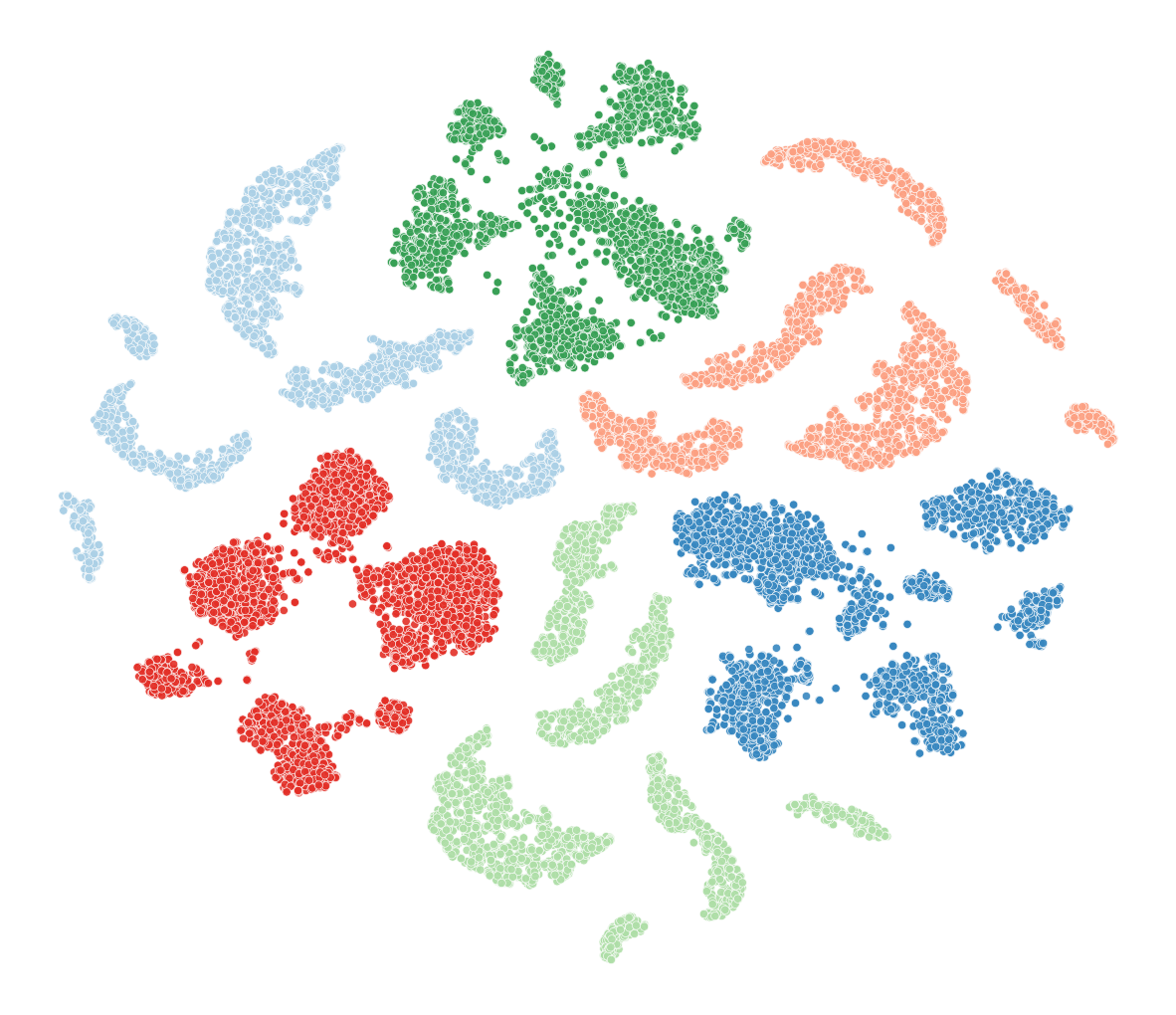}
        \caption{w/ Adversarial Training}
        \label{fig:top-right}
    \end{subfigure}
    \hspace{0.02\textwidth} 
    \begin{subfigure}[b]{0.2\textwidth}
        \centering
        \includegraphics[width=\textwidth]{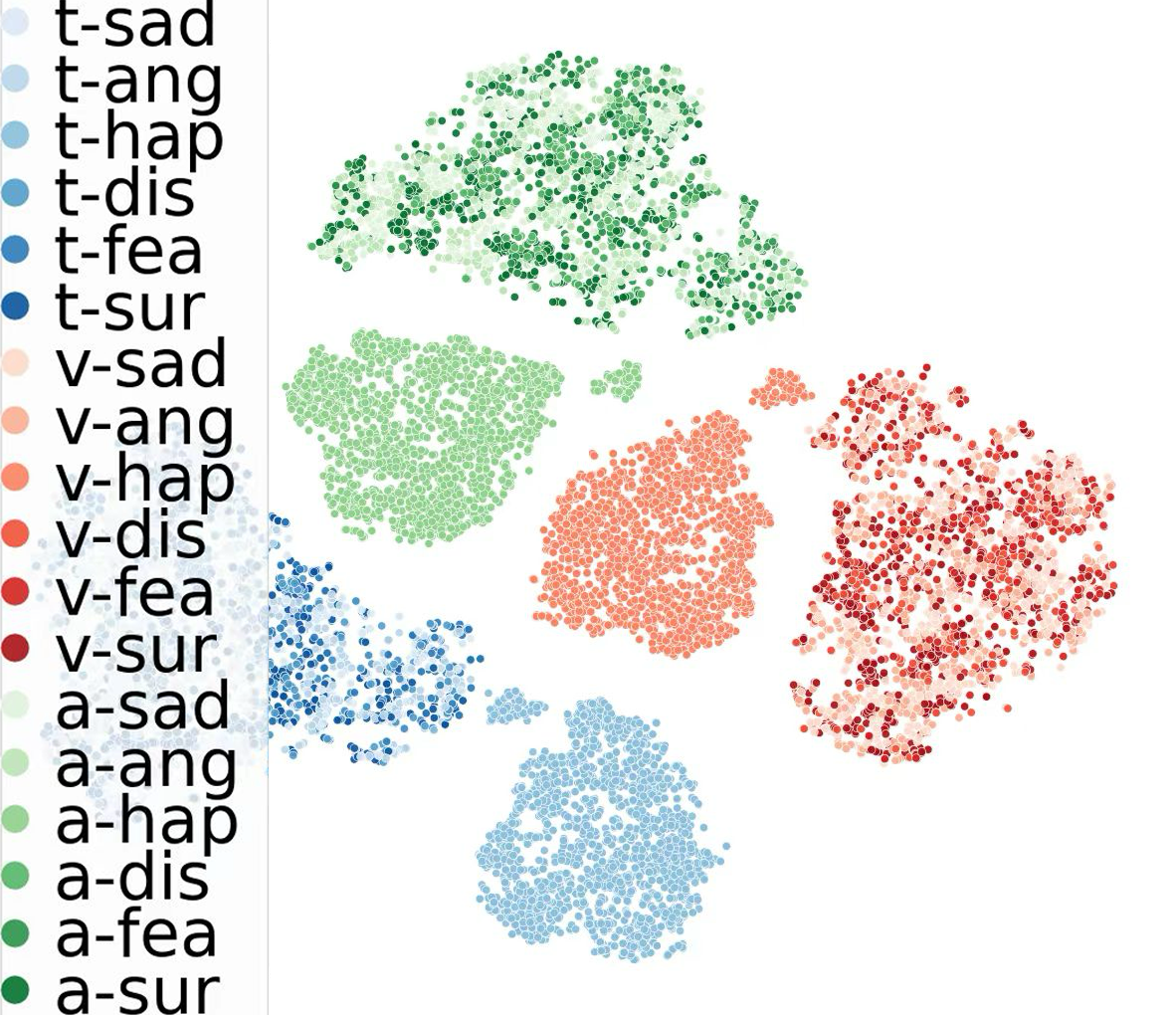}
        \caption{w/o RN and CLN}
        \label{fig:bottom-left}
    \end{subfigure}
    \hspace{0.02\textwidth}
    \begin{subfigure}[b]{0.2\textwidth}
        \centering
        \includegraphics[width=\textwidth]{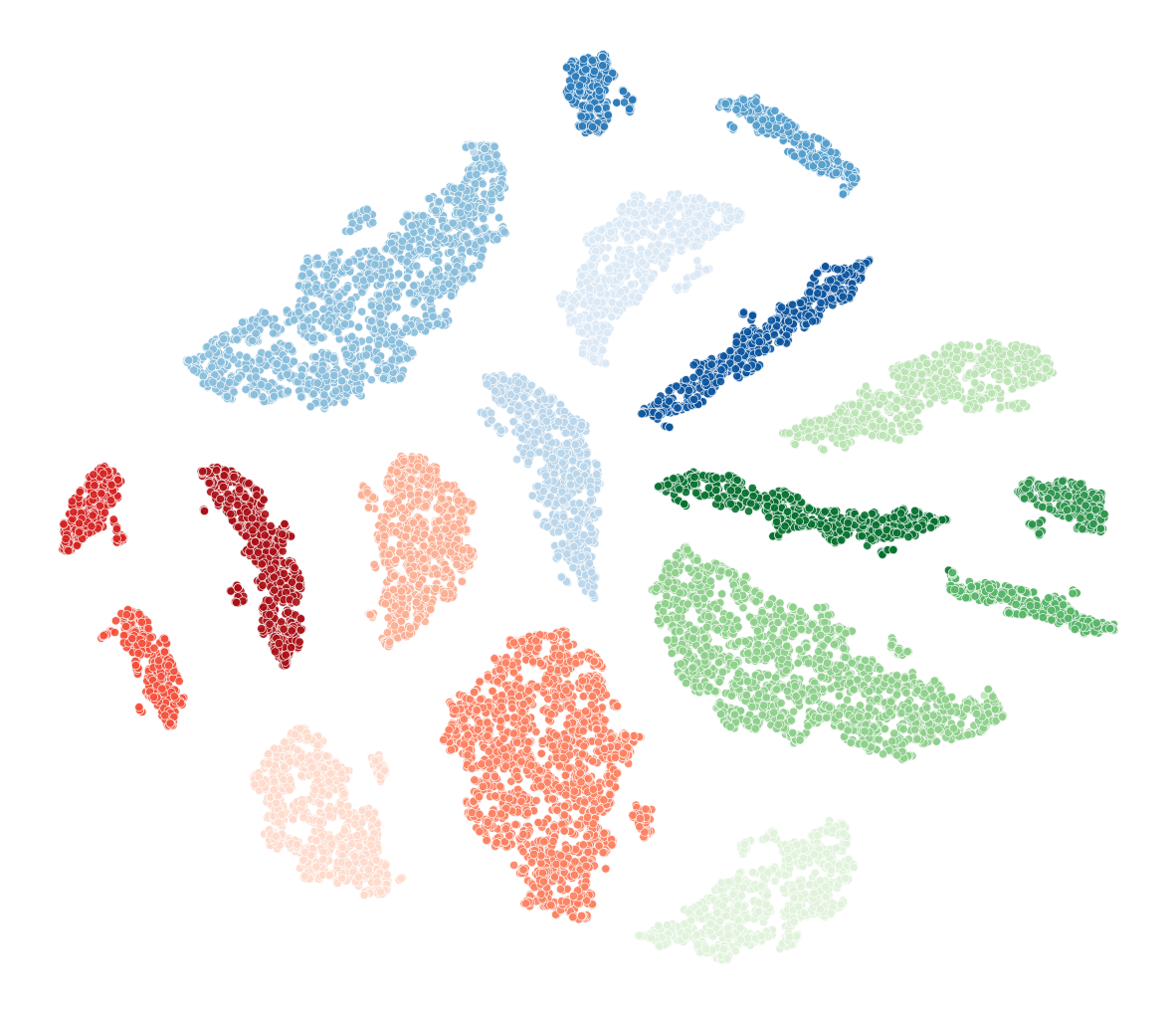}
        \caption{w/ RN and CLN}
        \label{fig:bottom_right}
    \end{subfigure}
    \vspace{-0.2em}
    \caption{t-SNE visualizations of modality embeddings. (a)(b): Specificity and commonality features without/with adversarial training. Color indicates modality (textual, visual, acoustic); saturation differentiates specificity (dark) and commonality (light) components. (c)(d): Reconstruction embeddings without/with RN and CLN, where hue denotes modality and saturation encodes emotion.}
    \label{fig:tsne}
\end{figure*}

\subsection{Comparison with the State-of-the-Art}
We present the performance comparisons of RAMer on the MEmoR, CMU-MOSEI, and M3ED datasets in Table~\ref{tab:primary}, Table~\ref{tab:cmu-mosei}, and Table~\ref{tab:m3ed}, respectively, with following observations.

1) On the MEmoR dataset, RAMer outperforms all baselines by a significant margin. While TAILOR achieves a high weighted-F1 score in the fine-grained setting, its overall performance is weaker due to biases toward frequent and easier-to-recognize classes. RAMer consistently delivers strong results across all settings, demonstrating its ability to learn more effective representations. 
2) On the CMU-MOSEI and M3ED datasets, RAMer surpasses state-of-the-art methods on all metrics except recall, which is less critical compared to accuracy and Micro-F1 in these contexts. 
3) Deep-based methods outperform classical ones, highlighting the importance of capturing label correlations for improved classification performance. 
4) Multimodal methods like HHMPN and AMP significantly outperform the unimodal ML-GCN, emphasizing the necessity of multimodal interactions.
5) Models optimized for dyadic conversations, such as CARAT, experience a notable performance drop in multi-party settings with incomplete modalities. In contrast, RAMer excels in both scenarios, achieving substantial improvements in Micro-F1 and Macro-F1 scores on the MEmoR dataset.

\subsection{Ablation Study}
To better understand the importance of each component of RAMer, we compared various ablated variants. 

As shown in Table ~\ref{tab:ablation}, we make the following observations:

\begin{itemize}
\item The specificity and commonality enhance MMER performance. Variants (1), (2), and (3) show lower Micro-F1 than variant (11). This indicates that jointly learning specificity and commonality yields superior performance, underscoring the importance of capturing both modality-specific specificity and shared commonality.

\item  Contrastive learning benefits the MMER. The inclusion of loss functions $\mathcal{L}_{scl}$ in adversarial training leads to progressive performance improvements, as evidenced by the superior results of (4). 


\item Feature reconstruction net benefits MMER. Variants (5), (6), (7) are worse than (11), and (8) shows an 0.045 decrease in Micro-F1, which indicates that feature reconstruction can improve model performance. When the entire reconstruction process is omitted, the performance of (8) declines even more compared to (6) and (7), confirming the effectiveness of multi-level feature reconstruction in achieving multi-modal fusion.

\item Changing the fusion order leads to poor performance, variants (9) and (10) perform worse than (11). It validates the rationality and optimality of feature fusion.
\end{itemize}

\begin{table}[t]
\begin{center} 
\resizebox{\columnwidth}{!}
{ 
\begin{tabular}{lcccc}
\hline
\hline
Approaches                                                                                   
& Acc   & P     & R     & Micro-F1 
\\ 
\hline


(1) w/o ${{L}_{sc}}$                                                                                      
& 0.474 & 0.610  & 0.517 & 0.573    
\\

(2) w/o $\bm{C}^{\{v,a,t\}}$                                                             
& 0.467 & 0.612 & 0.501 & 0.552    
\\

(3) w/o $\bm{S}^{\{v,a,t\}}$                                                            
& 0.460  & 0.599 & 0.491 & 0.552    
\\
\hline

(4) w/o ${{L}_{scl}}$                                                                               
& 0.492 & 0.651 & 0.540  & 0.588    
\\

(5) w/o $\varepsilon ^m$, $d^m$                                             
& 0.480  & 0.633 & 0.524 & 0.580     
\\

(6) w/o $\bm{\mathcal{X}}_\beta^m$                                              
& 0.481 & 0.641 & 0.538 & 0.590     
\\

(7) w/o $\bm{\mathcal{X}}_\gamma^m$                                               
& 0.485 & 0.620  & 0.523 & 0.586    
\\

(8) w/o $\bm{\mathcal{X}}_\beta^m$ + $\bm{\mathcal{X}}_\gamma^m$ 
& 0.477 & 0.603 & 0.490  & 0.557    
\\ 

\hline

(9) $\Lambda {\{v,t,a,\bm{C}\}}$                                                               
& 0.489 & 0.603 & 0.514 & 0.564    
\\ 

(10) $\Lambda {\{t,a,v,\bm{C}\}} $                                                               
& 0.494 & 0.650  & 0.525 & 0.582    
\\

\hline

(11) \textbf{RAMer}                                                                                       
&\textbf{0.502} &\textbf{0.672} & \textbf{0.545} & \textbf{0.602}   
\\ 
\hline
\hline
\end{tabular}
}
\vspace{-0.2em}
\caption{Ablation study on the aligned CMU-MOSEI dataset. $\Lambda$ refers to the fusion order, and ${{L}_{sc}}$ represents the specific and common loss. ``w/o $\varepsilon ^m$, $d^m$'' denotes the removal of the encoding and decoding processes.}
\label{tab:ablation} 
\end{center}
\vspace{-1.0em}
\end{table}

\subsection{Qualitative Analysis}
\subsubsection{Visualization of Learned Modality Representations}
To evaluate the effectiveness of reconstruction-based adversarial training, we use t-SNE to visualize the commonality and specificity representations learned in the aligned CMU-MOSEI dataset. In figure \ref{fig:tsne}(a), without adversarial training, specificity and commonality are loosely separated, but their distributions overlap in certain areas, such as the lower-right corner. In contrast, Figure~\ref{fig:tsne}(b) shows clearer separation of commonality and specificity, forming distinct boundaries and effectively distinguishing emotions across modalities. 
Figure \ref{fig:tsne}(c) demonstrates that without the reconstruction net (RN) and contrastive learning net (CLN), modality-wise embeddings are distinguishable, but emotion labels within the same modality remain intermixed. In contrast, Figure \ref{fig:tsne}(d) shows clear separation across both modalities and emotions, indicating that the reconstruction-based module improves representation distinctiveness. Overall, RAMer accurately captures both the commonality and specificity of different modalities.


\subsubsection{Visualization of Modality-to-Label Correlations}
\begin{figure}[t]
    \centering
\includegraphics[width=0.6\linewidth]{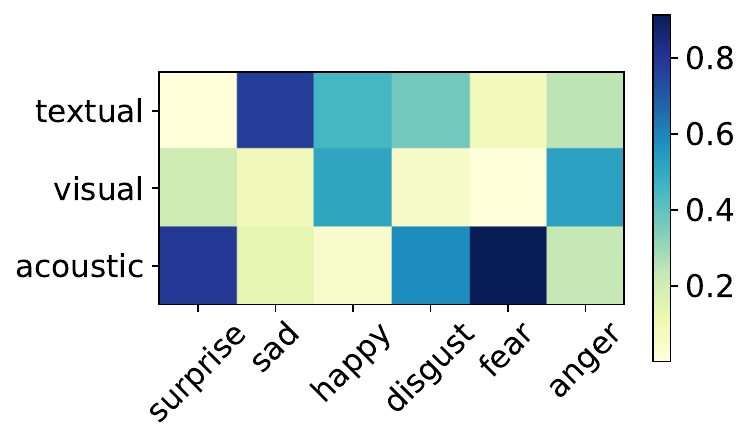}
    \caption{The correlation of modality-to-label dependencies.}
    \label{fig:modality-to-label}
\end{figure}

To explore the relationship between modalities and labels, we visualized the correlation of labels with their most relevant modalities. As shown in Figure \ref{fig:modality-to-label}, regardless of the presence of adversarial training, different emotion label is influenced by different modalities. For instance, surprise is predominantly correlated with the acoustic modality, while anger is primarily associated with the visual modality. This indicates that each modality captures the distinguishable semantic information of the labels from distinct perspectives.

\begin{figure}[t!] 
    \centering 
\includegraphics[width=0.98\linewidth]{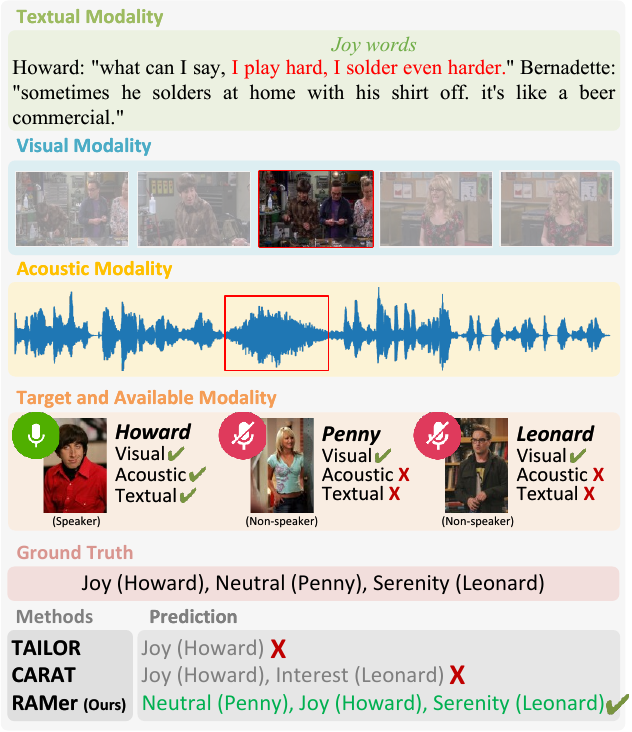}
    \caption{An example of the case study results.}
    \label{fig:case study}
\end{figure}

\subsubsection{Case Study}
To demonstrate RAMer’s robustness in complex scenarios, Figure~\ref{fig:case study} shows an example of MMER on the MEmoR dataset where specific target persons have incomplete modality signals. The top three rows display different modalities from a video clip, segmented semantically with aligned multi-modal signals. Key observations include: 1) The target moment requires recognizing emotions for both the speaker (e.g., Howard) and non-speakers (e.g., Penny and Leonard). While the speaker typically has complete multi-modal signals, non-speakers often lack certain modalities. TAILOR, limited by missing modalities, yields partial predictions as its self-attention mechanism struggles to align labels with missing features. 2) Limitations of single or incomplete modalities. A single modality, such as text, is often insufficient for accurate emotion inference (e.g., only Howard’s Joy is detectable from text alone). Although CARAT attempts to reconstruct missing information, it fails to capture cross-modal commonality, leading to incorrect predictions. 3)Inter-person interactions and external knowledge (e.g., personality traits) play important roles. Inter-person attention helps compensate for missing data, while personality-aware reasoning improves emotion inference across participants, highlighting the synergy between user profiling and emotion recognition. Experimental results demonstrate that RAMer achieves superior robustness and effectiveness in complex real-world scenarios.

%% file: source/conclusion.tex
\section{Conclusion}
In this paper, we proposed {RAMer}, a framework that refines multi-modal representations using reconstruction-based adversarial learning to address the Multi-party Multi-modal Multi-label Emotion Recognition problem. RAMer captures both the commonality and specificity across modalities using an adversarial learning module, with reconstruction and contrastive learning enhancing its ability to differentiate emotion labels, even with missing data. We also introduce a personality auxiliary task to complement incomplete modalities, improving emotion reasoning through modality-level attention. Furthermore, the stack shuffle strategy enriches the feature space and strengthens correlations between labels and modalities.
Extensive experiments on three datasets demonstrate that RAMer consistently outperforms state-of-the-art methods in both dyadic and multi-party MMER scenarios.

